\pdfoutput=1
\documentclass[11pt]{article}

\usepackage{acl}
\usepackage{times}
\usepackage{latexsym}
\usepackage{amsfonts}

\usepackage[T1]{fontenc}
\usepackage[utf8]{inputenc}

\usepackage{graphicx}
\usepackage{subcaption}

\usepackage{microtype}
\usepackage{cinzel} %inserted by steven to add MERLOT font

\usepackage{multirow}
\usepackage{tabularx}

\usepackage{comment} %note: inserted by Steven for block commenting
\usepackage{wrapfig} %note: inserted by Steven for wrapping text around figures
\usepackage{calligra}

\usepackage{aurical}

\newcommand\blfootnote[1]{%
  \begingroup
  \renewcommand\thefootnote{}\footnote{#1}%
  \addtocounter{footnote}{-1}%
  \endgroup
}

\makeatletter
\def\thickhline{%
  \noalign{\ifnum0=`}\fi\hrule \@height \thickarrayrulewidth \futurelet
   \reserved@a\@xthickhline}
\def\@xthickhline{\ifx\reserved@a\thickhline
               \vskip\doublerulesep
               \vskip-\thickarrayrulewidth
             \fi
      \ifnum0=`{\fi}}
\makeatother

\makeatletter
\newcommand\footnoteref[1]{\protected@xdef\@thefnmark{\ref{#1}}\@footnotemark}
\makeatother

\newlength{\thickarrayrulewidth}
\newcommand*{\affmark}[1][*]{\textsuperscript{#1}}
\newcommand{\DatasetName}{\textit{TT-Corp}}
\newcommand{\MethodName}{\textbf{\textcinzel{PANCETTA}}}

\newcommand{\MethodP}{\textbf{\textcinzel{PANCETTA-P}}}

\newcommand{\MethodJ}{\textbf{\textcinzel{PANCETTA-J}}}

\newcommand{\TaskPrompt}{TT-Prompt}
\newcommand{\TaskKeyword}{TT-Keyword}

\usepackage{amssymb}% http://ctan.org/pkg/amssymb
\usepackage{pifont}% http://ctan.org/pkg/pifont
\newcommand{\cmark}{\ding{51}}%
\newcommand{\xmark}{\ding{55}}%

\usepackage{tipa}
\usepackage{makecell}
\usepackage{enumitem,kantlipsum}

\newcommand{\titlecolor}{violet}
\title{\textcinzelblack{PANCETTA}: {\textcolor{\titlecolor}P}honeme
{\textcolor{\titlecolor} A}ware {\textcolor{\titlecolor}N}eural {\textcolor{\titlecolor}C}ompletion to {\textcolor{\titlecolor}E}licit {\textcolor{\titlecolor}T}ongue {\textcolor{\titlecolor}T}wisters {\textcolor{\titlecolor}A}utomatically}

\author{Sedrick Scott Keh \affmark[1], Steven Y. Feng\thanks{~~Work done while at CMU.}${}^{ }$ \bf{\thanks{\quad Equal contribution by Steven and Varun}}${}^{ }$ \affmark[2], Varun Gangal\footnotemark[2]${}^{ }$ \affmark[1], \\ 
\textbf{Malihe Alikhani \affmark[3], Eduard Hovy \affmark[1]} \\ 
\affmark[1]Carnegie Mellon University, \affmark[2]Stanford University, \affmark[3]University of Pittsburgh \\ 
\texttt{\{skeh,vgangal,hovy\}@cs.cmu.edu} \\ \texttt{syfeng@stanford.edu, malihe@pitt.edu}}

\begin{document}
\maketitle

\begin{abstract}
Tongue twisters are meaningful sentences that are difficult to pronounce. The process of automatically generating tongue twisters is challenging since the generated utterance must satisfy two conditions at once: phonetic difficulty and semantic meaning. Furthermore, phonetic difficulty is itself hard to characterize and is expressed in tongue twisters through a heterogeneous mix of phenomena such as alliteration and homophony. In this paper, we propose {\MethodName}: \textbf{P}honeme \textbf{A}ware \textbf{N}eural \textbf{C}ompletion to \textbf{E}licit \textbf{T}ongue \textbf{T}wisters \textbf{A}utomatically. We leverage phoneme representations to capture the notion of phonetic difficulty, and we train language models to generate original tongue twisters on two proposed task settings. To do this, we curate a dataset called {\DatasetName}, consisting of existing English tongue twisters. Through automatic and human evaluation, as well as qualitative analysis, we show that {\MethodName} generates novel, phonetically difficult, fluent, and semantically meaningful tongue twisters.
\end{abstract}

\section{Introduction}
\label{sec:intro}
A \emph{tongue twister} is a sentence which is both \emph{articulatorily difficult} (i.e. colloquially speaking, \emph{hard to say} or "twisting") 
while at the same time being \emph{meaningful and fluent}. Some examples of tongue twisters are shown in Table \ref{tab:tt-examples}.

Together with riddles, rhymes, fables, and other such creative artifacts, tongue twisters were historically often employed as a vehicle for early transmission of native language diction, grammar, and vocabulary to children, through parent-child interaction, playtime activity, and
kindergarten instruction \cite{akinyemi2003yoruba,10.1093/notesj/gjab168}. Tongue twisters have also been used as experimental
aids for research studies of speech production in cognitive science and related disciplines,
both among healthy speakers and those with speech and auditory disorders such as dysarthria \cite{kember2017inducing}. They are also used as pedagogic aids in speech therapy, as well as for treatment of speech disorders and psychological disorders relating to public speaking and elocution \cite{revathy2016enhancing}. An example of this was in a scene\footnote{\url{https://youtu.be/7WJts0gKCRM?t=53}} from \textit{The King's Speech (2010)}, where George VI repeats a tongue twister during therapy to reduce his stutter.
Lastly, tongue twisters find use as teaching aids for English  diction in EFL (English as a Foreign Language) instructional settings \cite{provsic2009use}.\blfootnote{Code can be found at: \url{https://github.com/sedrickkeh/PANCETTA}}

The coining of a novel, unique tongue twister which spreads sufficiently to become normative and well-recognized is rare, hence characterizing them as long-tailed linguistic phenomena \cite{naik2022domain}.
However, they are not limited to English, and are found across the world's languages, e.g., Persian (\textit{"Shish sikh jigar sikhi shi shezar."}) \cite{jam2018features} and French (\textit{"Cinq chiens chassent six chats."})

\begin{figure}
    \centering
    \includegraphics[width=0.48\textwidth]{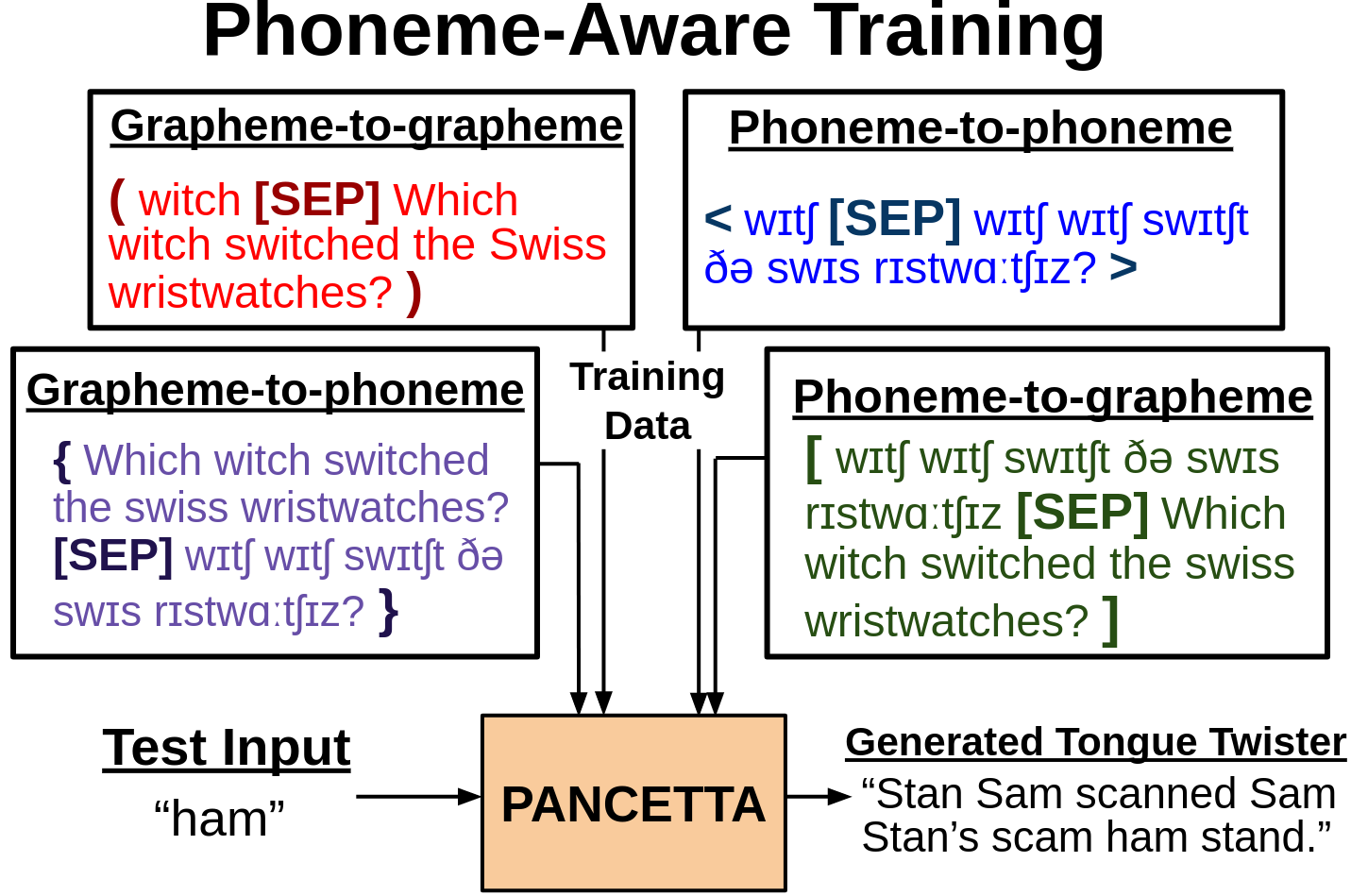}
    \caption{Overview of the phoneme-aware training in the {\MethodName} model.}
    \label{fig:main-diagram}
\end{figure}

\begin{table*}[t]
\footnotesize
\centering
        \begin{tabular}{c|p{0.10\textwidth}p{0.38\textwidth}p{0.28\textwidth}}
            \hline
            \textbf{Task} & \textbf{Input} & \textbf{Tongue Twister} & \textbf{Tongue Twister (Phoneme)} \\
            \hline \hline
            \multirow{3}{*}{TT-Prompt} & \multicolumn{1}{p{0.10\textwidth}}{A good cook} & \multicolumn{1}{p{0.35\textwidth}}{A good cook could cook as many cookies as a good cook who could cook cookies.} & \multicolumn{1}{p{0.28\textwidth}}{\textipa{@ gUd kUk kUd kUk æz mEni kUki:z æz @ gUd kUk hu: kUd kUk kUki:z.}} \\\cline{2-4}
                                 & \multicolumn{1}{p{0.08\textwidth}}{Chubby jugglers} & \multicolumn{1}{p{0.35\textwidth}}{Chubby jugglers juggling oranges jovially.} & \multicolumn{1}{p{0.28\textwidth}}{\textipa{tS2bi dZ2g@lRz dZ2g@lIN Or@ndZ@z dZoUveIli}.} \\\cline{2-4}
                                & \multicolumn{1}{p{0.10\textwidth}}{Does the} & \multicolumn{1}{p{0.33\textwidth}}{Does the rapid rabid rabbit wrap it?} & \multicolumn{1}{p{0.28\textwidth}}{\textipa{d2z D@ ræp@d ræbId ræb@t ræp It?}} \\\hline
            \multirow{3}{*}{TT-Keyword} & \multicolumn{1}{p{0.10\textwidth}}{shoes, dog} & \multicolumn{1}{p{0.39\textwidth}}{If a dog chews shoes, whose shoes does he choose?} & \multicolumn{1}{p{0.32\textwidth}}{\textipa{If @ dOg tSu:z Su:z, hu:z Su:z d2z hi: tSu:z}?} \\\cline{2-4}
                                 & \multicolumn{1}{p{0.10\textwidth}}{blood, death} & \multicolumn{1}{p{0.35\textwidth}}{Bad dead bed-bugs bleed bug blood.} & \multicolumn{1}{p{0.28\textwidth}}{\textipa{bæd dEd bEd-b2gz bli:d b2g bl2d.}} \\\cline{2-4}
                                & \multicolumn{1}{p{0.07\textwidth}}{king, art, wall} & \multicolumn{1}{p{0.35\textwidth}}{A truly rural frugal ruler's mural was on the wall.} & \multicolumn{1}{p{0.28\textwidth}}{\textipa{@ tru:li rUr@l fru:g@l ru:lRz mjUr@l wA:z A:n D@ wOl.}} \\\hline
        \end{tabular}
    \caption{Example inputs and target outputs for both the TT-Prompt and TT-Keyword task settings, along with the phoneme representations of the tongue twisters.}
    \label{tab:tt-examples}
\end{table*}

Consider the idealized scenario where we have a \textit{"mouth model"} which a) maps different regions of the mouth, palate, and the larynx to the dictionary of fundamental sounds, i.e. phones being produced, and b) based on this grounding, can quantify the hardness of producing one sound after another by inducing a distance measure between any phone pair. Assuming access to this idealized model, one could deconstruct the process of generating a tongue twister as sampling a sequence of preferably difficult/distant phone-phone transitions starting with an initial sequence of one or more phones (which could come from a prompt, or be chosen uniformly, based on the task setting).

However, there are several impediments which make realizing such an idealized model considerably intractable. First, the dictionary of fundamental sounds at the granularity we use in practice, i.e. at the level of phonemes, does not neatly map to particular points of the palate \cite{ladefoged2014course}.
Rather, each phoneme itself corresponds to a set of actions involving multiple organs and palatal regions. For instance, velar consonants like \textit{k} are produced based on tongue-velum (soft upper palate) interaction. Secondly, a tongue twister as per its definition is not merely a difficult to pronounce sequence of phonemes, but also one that maps to a meaningful and fluent sequence of words. How one can maintain this property in conjunction with the process of sampling difficult transitions from the mouth model's space is unclear.

The automatic generation of tongue twisters has largely been unexplored. This task is challenging because it requires being able to model phonetic difficulty of various syllables and tokens, which is not something that existing language models are trained to do. To achieve this, we have to work in the phoneme space. Phonemes have previously been used to aid in speech recognition \cite{sundararaman2021phoneme} and rhyme generation \cite{hopkins-kiela-2017-automatically}. 
We hypothesize that by working with phonemes, we will be able to model and generate patterns that characterize phonetic difficulty.

Tongue twisters go beyond relatively simpler phonetic phenomena such as alliteration, since they employ a heterogeneous mix of strategies \cite{jorgensen1981tickled} including alliteration itself (\textit{\textbf{s}he \textbf{s}ells \textbf{s}eashells}), use of homophonic words/subwords (\textit{sells}/\textit{-shells}, \textit{she}/\textit{sea-}), and alternating between similar start phonemes for tokens (\textit{s} and \textit{sh}), sometimes even using multiple such phenomena in conjunction within the same example to create the cumulative effect of articulatory difficulty.

Our contributions are as follows: (1) We curate a dataset, {\DatasetName}, of diverse tongue twisters. (2) We present two new task settings ({\TaskPrompt} and {\TaskKeyword}) for automatic tongue twister generation, and we design and evaluate simple baselines for these tasks. (3) We propose a phoneme-aware method called {\MethodName}, which models and generates coherent and phonetically difficult phrases by taking phonemes into account. We show that {\MethodName} generates higher-quality tongue twisters through both automatic and human evaluations and qualitative analysis of the outputs.

\section{Task Settings and Dataset}
\label{sec:dataset}
\subsection{Task Settings}
We propose two settings for automatic tongue twister generation. We call these tasks {\TaskPrompt} and {\TaskKeyword}. Examples of these two tasks can be found in Table \ref{tab:tt-examples}, and they are detailed below:
\begin{enumerate}[wide, labelwidth=!, labelindent=0pt]
    \item \textbf{Generating tongue twisters from prompts ({\TaskPrompt})}: Given a few words to start a sentence, the goal is to complete the sentence in a coherent way such that the resulting generation is a tongue twister. Prompts can be of varying lengths.
    \item \textbf{Generating tongue twisters from keywords ({\TaskKeyword})}: Given a set of keywords, the goal of this task is to generate a coherent tongue twister which incorporates the semantics of the keywords. The set of keywords can be of varying sizes. These keywords do not necessarily have to appear verbatim and do not necessarily have to appear in order.
\end{enumerate}

\subsection{{\DatasetName} Dataset}
\label{subsec:dataset}
As previously noted, tongue twisters are long-tailed linguistic phenomena, and it is rare to find new unique tongue twisters. Given this, we curate a dataset of 644 unique English tongue twisters into a dataset called {\DatasetName}. These tongue twisters are compiled from various sources, ranging from blog posts to English learning websites. A more detailed list of these sources and data processing details can be found in Appendix \ref{appendix:dataset-details}. Despite the seemingly small scale of this dataset, multiple studies have successfully generated creative text even when training data is limited: 511 personifications \cite{keh-etal-2022-pineapple}, 1400 MadLibs \cite{hossain-etal-2017-filling}, 401 portmanteaus \cite{deri-knight-2015-make}, and 576 clippings \cite{mattiello2013}.

We also create a non-tongue twister version of each input in {\DatasetName}, which will later be used to explore style transfer models (\S\ref{sec:models}) and to train a phonetic difficulty classifier (\S\ref{sec:eval-auto}). This is done through synonym replacement. First, we determine the parts-of-speech of all the words in the sentence and identify the nouns, verbs, and adjectives.\footnote{The spaCy library \cite{spacy2} was used to extract the POS tags.} We then use WordNet \cite{wordnet} to generate a list of synonyms for each of these nouns, verbs, and adjectives, and we select the highest ranked replacement which shares the same part-of-speech. Examples of this process are shown in Table \ref{tab:non-tt-examples}.

\begin{table}[t]
    \footnotesize
    \centering
    \begin{tabular}{p{0.45\columnwidth}|p{0.46\columnwidth}}
        \hline
        \textbf{Tongue Twister} & \textbf{Non-TT Version} \\
        \hline \hline
        There was a \textcolor{red}{little} \textcolor{blue}{witch} which \textcolor{orange}{switched} from Chichester to Ipswich. (ex.1) & There was a \textcolor{red}{small} \textcolor{blue}{enchantress} which \textcolor{orange}{exchanged} from Chichester to Ipswich. \\
        \hline
        He \textcolor{red}{wanted} to \textcolor{blue}{desert} his \textcolor{orange}{dessert} in the desert. (ex.2) & He \textcolor{red}{desired} to \textcolor{blue}{abandon} his \textcolor{orange}{sweet} in the desert. \\
        \hline
        \textcolor{red}{Tie} a \textcolor{blue}{tight} \textcolor{orange}{knot} in the \textcolor{olive}{shape} of a \textcolor{violet}{nought}. (ex.3) & \textcolor{red}{Bind} a \textcolor{blue}{taut} \textcolor{orange}{gnarl} in the \textcolor{olive}{form} of a \textcolor{violet}{zero}. \\
        \hline
    \end{tabular}
    \caption{Examples of the synonym replacement process to generate non-tongue twister versions of the tongue twisters in {\DatasetName}. Different colors are used to indicate which words are replaced by their synonyms.}
    \label{tab:non-tt-examples}
\end{table}

One key advantage of this synonym replacement process is that it can replace a word according to its part-of-speech in the sentence. In the second example in Table \ref{tab:non-tt-examples}, the word "desert" appears twice -- first as a verb (which is replaced with "abandon"), and again as a noun (which is not replaced). However, this synonym replacement process does not take the context of the words into account. In the third example, while the individual synonym replacements make sense on their own, the final sentence sounds quite unnatural. 
For our purposes, however, this is not a significant issue: we do not need the replacement sentences to be absolutely perfect, as the quality of the ground-truth tongue twister is more important. This will be explained further when we use this parallel dataset in \S\ref{sec:eval-auto}.

\section {Methodology}
\label{sec:methodology}
As this is a new task, there are no existing methods that can easily generate novel tongue twisters. The main challenge is how to incorporate phonetic difficulty into our generations. To do so, we propose two baseline and two phoneme-aware models, which are applicable to both the {\TaskPrompt} and {\TaskKeyword} task settings (see Table \ref{tab:model-summary}).

\begin{table*}[t]
    \footnotesize
    \centering
    \begin{tabular}{c|c|c|c|c}
         \textbf{Method Name} & \textbf{Models Used} & \textbf{Description} & \makecell{Phoneme \\ Representation} & Leverage Pretraining \\
         \hline
         Grapheme-based Methods & GPT-2, GPT-J & g2g & \xmark & \cmark \\
         Style Transfer Methods & BART, T5 & g2g + g2g & \xmark & \cmark \\
         {\MethodP} & GPT-2, BART & g2p + p2p + p2g & \cmark & \xmark \\
         {\MethodJ} & GPT-2, GPT-J & \makecell{g2g, p2p, g2p, p2g \\ (only g2g during test-time)} & \cmark & \cmark \\
    \end{tabular}
    \begin{tabular}{c|p{0.75\textwidth}}
        \textbf{Method Name} & \textbf{Example} \\
        \hline
        Grapheme-based Methods & She sells $\rightarrow$ She sells seashells on the seashore.\\
        Style Transfer Methods & She sells $\rightarrow$ She sells things on the beach. $\rightarrow$ She sells seashells on the seashore.\\
        {\MethodP} & She sells $\rightarrow$ \textipa{Si: sElz} $\rightarrow$ \textipa{Si: sElz si:SElz A:n D@ si:SOr} $\rightarrow$ She sells seashells on the seashore.\\
        {\MethodJ} & She sells $\rightarrow$ She sells seashells on the seashore. \\
        & She sells seashells on the seashore. $\rightarrow$ \textipa{Si: sElz si:SElz A:n D@ si:SOr.} \\
        & \textipa{Si: sElz si:SElz A:n D@ si:SOr.} $\rightarrow$ She sells seashells on the seashore. \\
        & \textipa{Si: sElz} $\rightarrow$ \textipa{Si: sElz si:SElz A:n D@ si:SOr.}\\
    \end{tabular}
    \caption{Summary of the models discussed in \S\ref{sec:models}, along with some examples.}
    \label{tab:model-summary}
\end{table*}

\subsection{Models}
\label{sec:models}
\begin{enumerate}[wide, labelwidth=!, labelindent=0pt]
    \item \textbf{Grapheme-based Methods (g2g)} -- We treat tongue twister generation as a seq2seq task, where the prompt (for {\TaskPrompt}) or keywords (for {\TaskKeyword}) is the input, and the tongue twister is the target output. 
    We fine-tune GPT-2 \cite{radford2019language} and GPT-J \cite{gpt-j} using the input sequences "(X [SEP] Y)", where X represents the prompt/keywords, Y represents the tongue twister, and [SEP] is a separator token.
    
    \item \textbf{Style Transfer Methods} -- Given a prompt or a set of keywords, we generate a sentence (not necessarily a tongue twister) using GPT-2. This is easy for {\TaskPrompt} since GPT-2 is trained to do causal language modeling. For {\TaskKeyword}, we need to first train a GPT-2 model to perform keyword-to-text. We sample 10,000 sentences from WikiText-103 \cite{merity2016pointer} and extract their keywords using KeyBERT \cite{grootendorst2020keybert}. We then fine-tune GPT-2 using an "(X [SEP] Y)" template as described in the g2g methods. Here, X represents the keywords, and Y represents the corresponding sentence.
    
    We then attempt to convert these generated natural sentences into tongue twisters. We treat this as a seq2seq task and train a seq2seq model using our parallel dataset. During training, we use the non-TT versions as inputs and the tongue twisters as the ground truth target outputs. We use BART \cite{lewis-etal-2020-bart} and T5 \cite{2020t5} models for this seq2seq task. 

    \begin{figure}
        \centering
        \includegraphics[width=0.45\textwidth]{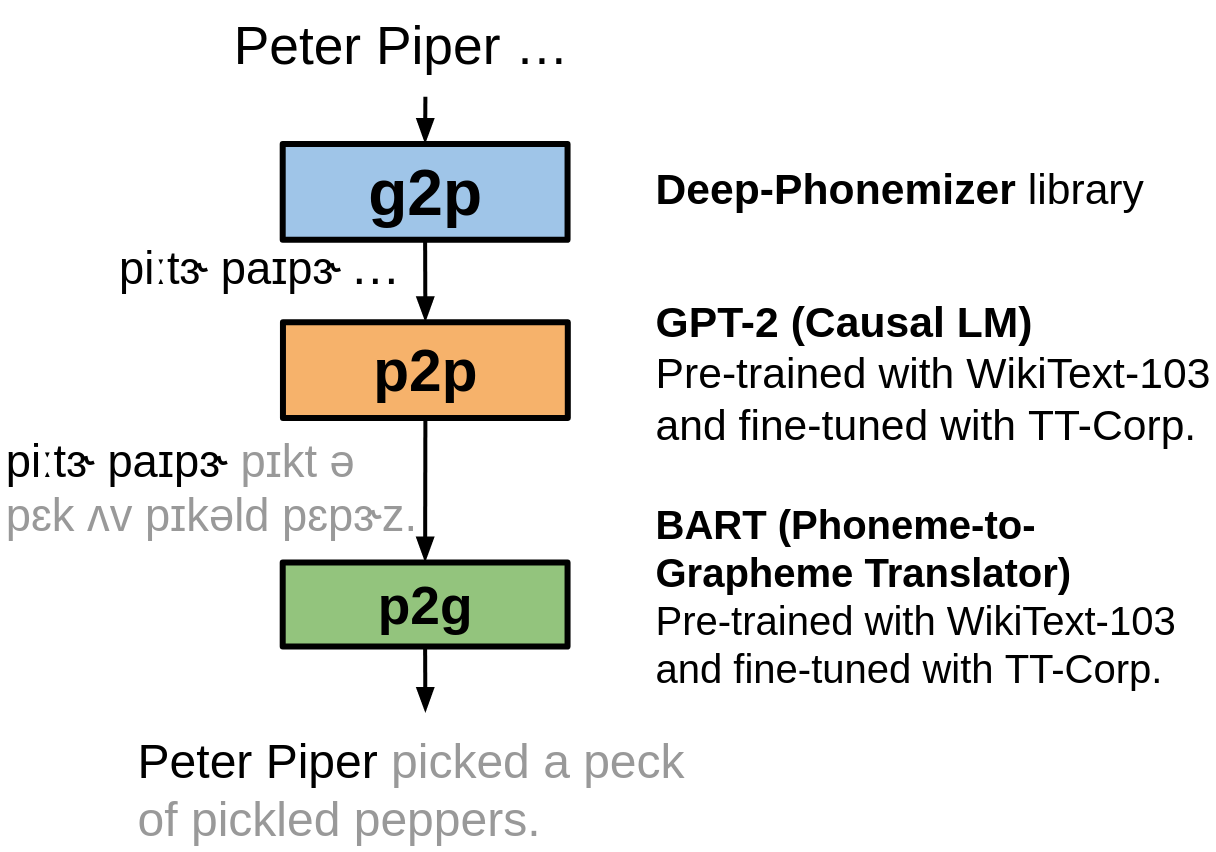}
        \caption{Overview of {\MethodP} pipeline.}
        \label{fig:pancetta-p}
    \end{figure}
    \item {\MethodP} 
    (Phoneme) -- For the previous g2g models, the fine-tuning was done only using graphemes. Because graphemes are not always representative of pronunciation, we hypothesize that it may be difficult for such models to capture information regarding the pronunciation. If we instead had a generative model which works on the phoneme space, then we could fine-tune this model on the tongue twister phonemes and hope that it can better capture these phonetic cues.
    
    We first pretrain a GPT-2 model to perform causal LM generation for phonemes. Pretraining is done using WikiText: we first convert all the WikiText sentences into their IPA phoneme representations and train a GPT-2 model on it.\footnote{The \texttt{deep-phonemizer} Python package was used for g2p transliteration.} For {\TaskKeyword}, instead of training a causal phoneme LM, we train to generate from keywords, using the (X [SEP] Y) template previously described. While there are multiple g2p phonemization toolkits, there are no readily available p2g toolkits that work well. Hence, we train our own p2g model. We treat this as a seq2seq translation task, once again using WikiText. We train a BART model with the phonemes as the inputs and the graphemes as the targets. Once both the p2p generation and the p2g translation models are trained, we then fine-tune the p2p models on the tongue twister phonemes (all steps similar to g2g), then use the p2g model to retrieve the grapheme representation of the outputs (see Figure \ref{fig:pancetta-p}). Lastly, since there is no capitalization in the phoneme space, we have to fix the capitalization of the generated outputs. We use the FastPunct library for this.\footnote{https://github.com/notAI-tech/fastPunct} Unlike the previous g2g methods, we only use GPT-2 (and not GPT-J) for {\MethodP} because GPT-J is too large to pretrain in a reasonable fashion.
    
    \item {\MethodJ} (Joint) -- One drawback of {\MethodP} is that because we do our own pre-training on the phoneme space, we are not able to leverage the existing pre-training of large language models. In order to leverage both the phoneme representations and the pre-training of GPT, we propose {\MethodJ}. This is similar to g2g, but instead of only training with the template (X [SEP] Y), we train with 4 different templates, representing 4 different modalities. Specifically, we train with the templates ($PK_G$ [SEP] $TT_G$), <$PK_P$ [SEP] $TT_P$>, [$TT_G$ [SEP] $TT_P$], and \{$TT_P$ [SEP] $TT_G$\}, where $PK$ represents the prompts/keywords, $TT$ represents the tongue twisters, $G$ represents the grapheme representation, and $P$ represents the phoneme representation. These 4 modalities represent g2g, p2p, g2p, and p2g respectively. Here, the type of surrounding brackets function similar to custom tokens which indicate the modality for the model.
    
    At test time, the model can be directly decoded using g2g mode without requiring phoneme information. We hypothesize that the phonetic structures learned during training time can serve as an effective "scaffold" --- being used explicitly only during finetuning time \cite{swayamdipta2018syntactic}.
\end{enumerate}

\subsection{Evaluation Metrics}
\subsubsection{Automatic Evaluation}
\label{sec:eval-auto}
As described in \S\ref{sec:intro}, a good tongue twister needs to satisfy two criteria: it needs to be both difficult to pronounce, as well as semantically coherent. We consider these two notions separately.
\begin{enumerate}[wide, labelwidth=!, labelindent=0pt]
    \item \textbf{Phonetic Difficulty}: We fine-tune a pretrained BERT-base classifier to differentiate between tongue twisters and regular sentences. To train this model, we use the parallel dataset of (TT, non-TT) pairs as described in \S\ref{subsec:dataset}. We specifically use these (TT, non-TT) pairs so that the model learns to classify based on phonetic difficulty rather than semantics. However, as mentioned in \S\ref{subsec:dataset}, some replacement sentences may sound unnatural. To ensure that the classifier learns to differentiate tongue twisters instead of picking up on these false signals, we augment our dataset with additional negative examples consisting of 500 sentences randomly sampled from WikiText.
    Rather than directly training on the sentences, we first convert the sentences to phoneme representations and train a classifier on the phonemes. 
    This trained BERT classifier achieves an 83.4 F1-score on the test set of the parallel dataset, indicating that it is indeed successful at discriminating phonetically easy and difficult sentences. To further verify that this metric successfully measures phonetic difficulty, we show that it correlates well with human-annotated measures of phonetic difficulty (\S \ref{sec:human_eval_results} and Table \ref{tab:correlations}).
    
    \item \textbf{Fluency:} We not only want phonetically difficult sentences; they must also be fluent and coherent. To measure this, we use the generation (log-perplexity) losses from a pretrained GPT-2.
    
    \item \textbf{Keyword Relevance} (only for {\TaskKeyword}): In {\TaskKeyword}, we want to ensure that the generated tongue twister is semantically similar to the keywords used. To measure this, we use the BERT embedding of keywords and compare it with the embedding of the target sentence. More specifically, we use the BERTScore \cite{bert-score} between the generated sentence and the "sentence" consisting of the keywords separated by commas.
\end{enumerate}

% We do not use metrics based on reference matching with the reference/gold outputs. This is because tongue twister generation from prompts/keywords is inherently a creative task with high output diversity and a large subspace of valid outputs.  

\subsubsection{Human Evaluation}
\label{sec:eval-human}
The human evaluation metrics are very similar to the ones in \S\ref{sec:eval-auto}. These are as follows: (1) \textbf{Phonetic Difficulty} ("\textit{How hard is the sentence to pronounce? To get a better sense of the difficulty, try saying the sentence out loud, quickly, and multiple times.}") and (2) \textbf{Fluency} ("\textit{Does it sound like good English with good grammar?}") Evaluations were done on a scale of 1-5, with 5 being the highest. Further details about human evaluation are in \S\ref{subsec:human-eval}.

\section{Experimental Setup}
\label{sec:experimental_setup}
\subsection{Implementation Settings}
\label{sec:implementation-settings}

\textbf{Prompt / Keyword Extraction:} To extract prompts for {\TaskPrompt}, we simply consider the first three words of each sentence by checking for the whitespace character. To extract keywords for {\TaskKeyword}, we use the KeyBERT library \cite{grootendorst2020keybert}, which returns keywords ranked by their cosine similarity scores to the entire sentence itself.
For each sentence, we consider the top 5 keywords as our set of keywords. When a sentence has <5 keywords, we simply take all the keywords. In our dataset, 39.56\% of the examples had <5 keywords.

\textbf{Dataset splits:} We split {\DatasetName} into a training-validation-test split with a 70-15-15 ratio. We use the same splits across all models and across both {\TaskPrompt} and {\TaskKeyword} task settings, as well as for training the phonetic difficulty classifier.

\textbf{GPT-2 fine-tuning} (g2g, {\MethodP}, {\MethodJ}): We use the pretrained GPT2-base (124M params.) and fine-tune for 5 epochs with a learning rate of 5e-5 and 100 warmup steps. 

\textbf{BART pretraining} ({\MethodP}): For the p2g model, we pretrain BART-base (139M params.) on WikiText-103 phonemes. We split WikiText into train-validation-test splits of 80-10-10. The final training set has size 523k. Training was done for 20 epochs with batch size of 16, learning rate of 5e-4, and weight decay of 0.1 with a cosine scheduler.

\begin{table*}[t]
    \scriptsize
    \centering
    \begin{tabular}{c|cc|ccc}
        & \multicolumn{2}{c}{\textbf{{\TaskPrompt}}} & \multicolumn{3}{c}{\textbf{{\TaskKeyword}}} \\
         \textbf{Method} & \textbf{Phon. Difficulty} & \textbf{Fluency} $\downarrow$ & \textbf{Phon. Difficulty} & \textbf{Fluency} $\downarrow$ & \textbf{Keyword Relevance} \\
         \hline \hline
         g2g (GPT-2) & 0.774 & 5.433 & 0.786 & 5.224 & 0.795 \\
         g2g (GPT-J) & 0.848 & 5.643 & 0.856 & 5.593 & 0.794 \\
         \hline
         Style Transfer (GPT-2+BART) & 0.672 & 4.356 & 0.472 & \textbf{3.662} & 0.783\\
         Style Transfer (GPT-2+T5) & 0.631 & \textbf{4.256} & 0.414 & 4.309 & 0.780 \\
         \hline
         {\MethodP} (GPT-2+BART) & 0.794 & 5.986 & 0.871 & 6.596 & 0.801 \\
         \hline
         {\MethodJ} (GPT-2) & 0.785 & 5.244 & 0.803 & 5.058 & \textbf{0.803} \\
         {\MethodJ} (GPT-J) & \textbf{0.866} & 5.718 & \textbf{0.888} & 5.169 & 0.800 \\
         \hline \hline
         Gold Outputs & 0.925 & 5.745 & 0.925 & 5.745 & 0.812 \\
    \end{tabular}
    \caption{Automatic evaluation averages for both {\TaskPrompt} and {\TaskKeyword}. The best-scoring method for each metric is highlighted in \textbf{bold}. Higher scores are better for all metrics except for fluency.}
    \label{tab:results-auto-full}
\end{table*}

\textbf{BART \& T5 fine-tuning} (Style Transfer): We fine-tune BART-large (406M params.) \& T5-large (737M params.) for 30 epochs with a batch size of 16, learning rate of 2e-5, and 400 warmup steps.

\textbf{GPT-J fine-tuning} (g2g, {\MethodJ}): Because GPT-J is too large (6B parameters), we use a compressed version of GPT-J, \footnote{https://huggingface.co/hivemind/gpt-j-6B-8bit} which incorporates various techniques such as 8-bit quantization \cite{dettmers2021optim8bit} and low-rank adaptation \cite{hu2021lora}. To fine-tune, we use 10 epochs, a batch size of 1, and a learning rate of 1e-5.

For the GPT models (i.e. GPT-2 and GPT-J), generation was done using nucleus sampling (p=1.0, temperature=0.8). Meanwhile, for the BART and T5 models, generation was done using beam search with a beam size of 5.
Additional hyperparameters and details on implementation can be found in Appendix \ref{appendix:hyperparameter-settings}. 

\subsection{Human Evaluation Settings}
\label{subsec:human-eval}
Human evaluation was done on Amazon Mechanical Turk (AMT). We selected annotators with >97\% HIT approval rate from Anglophone countries. \footnote{More details about the human eval are in Appendix \ref{appendix:human-eval}.} In each HIT, we present the generated outputs for each example in randomized order, and each test example was evaluated by exactly 2 annotators. 

We conduct two rounds of annotation, one for {\TaskPrompt} and another for {\TaskKeyword}. Within each round, we further subdivide annotating GPT-2 experiments and GPT-J experiments. This GPT-2/GPT-J split only applies to g2g and {\MethodJ} models; for the style-transfer and {\MethodP} models, we keep the same models for both rounds of evaluation. This is done to ensure that we only have one independent variable so that the changes in performance are due to the methodologies rather than the size of the models. Another reason for subdividing GPT-2/GPT-J experiments is so that we do not subject annotators to information overload from having to annotate too many similar examples. Owing to the same consideration, we also decided to omit human evaluation on the Style Transfer T5 baseline because we found it very similar to Style Transfer BART.

\section{Results and Analysis}
\label{sec:results_and_analysis}
\begin{table*}[t]
    \scriptsize
    \centering
    \begin{tabular}{c|cc|cc}
         & \multicolumn{2}{c}{\textbf{{\TaskPrompt}}} & \multicolumn{2}{c}{\textbf{{\TaskKeyword}}} \\
         \textbf{Method} & \textbf{Phonetic Difficulty} & \textbf{Fluency} & \textbf{Phonetic Difficulty} & \textbf{Fluency} \\
         \hline
         g2g (GPT-2) & 3.056 & 3.736 & 3.847 & \textbf{4.139} \\
         Style Transfer (GPT-2+BART) & 2.569 & 3.639 & 3.500 & 3.819 \\
         {\MethodP} (GPT-2+BART) & \textbf{3.528} & \textbf{3.778} & 3.722 & 3.764 \\
         {\MethodJ} (GPT-2) & 3.153 & 3.764 & \textbf{3.889} & 3.931 \\
         Gold Outputs & 3.361 & 3.931 & 3.833 & 4.000 \\
        \hline \hline
         g2g (GPT-J) & 3.521 & \textbf{3.979} & 3.791 & 3.708 \\
         Style Transfer (GPT-2+BART) & 3.271 & 3.75 & 3.25 & 3.750 \\
         {\MethodP} (GPT-2+BART) & \textbf{3.854} & 3.708 & 3.833 & \textbf{3.896} \\
         {\MethodJ} (GPT-J) & 3.708 & 3.646 & \textbf{3.979} & 3.604 \\
         Gold Outputs & 3.750 & 4.000 & 4.104 & 3.729 \\
    \end{tabular}
    \caption{Human evaluation averages for {\TaskPrompt} and {\TaskKeyword}. The top half of the table correspond to methods using GPT-2 as the base architecture, while the bottom half of the table correspond to methods using GPT-J as the base architecture. Top method scores for each metric are highlighted in \textbf{bold}.}
    \label{tab:results-human-full}
\end{table*}

\subsection{Automatic Evaluation Results}
\label{subsec:auto-eval-results}
Table \ref{tab:results-auto-full} shows the average results for the metrics outlined in \S\ref{sec:eval-auto}. From the phonetic difficulty results, we see that our proposed {\MethodName} models score higher than the baselines. More specifically, comparing g2g (GPT-2) (0.774) vs {\MethodJ} (GPT-2) (0.785) and comparing g2g (GPT-J) (0.848) vs {\MethodJ} (GPT-J) (0.866), we see that incorporating phoneme representations indeed aids in producing more phonetically difficult sentences. This pattern also holds true for {\TaskKeyword}, where both GPT-2 and GPT-J see increases in performance after incorporating phonemes. We also observe that {\MethodP} performs reasonably well in phonetic difficulty and has the highest score of the non-GPT-J models for both {\TaskPrompt} and {\TaskKeyword}. In fact, for {\TaskKeyword}, {\MethodP} is able to get very close to {\MethodJ} (GPT-J), which is remarkable, considering that it is only using a GPT-2 model. 

For fluency, style transfer models score better than the other models. This is likely because style transfer models first generate a regular sentence, then attempt to \textit{"tongue twisterize"} it. However, there is no guarantee that the sentence can even be reasonably converted to a tongue twister, resulting in minimal changes being made to the original GPT-generated sentence, thereby leading to good fluency scores when using perplexity. Meanwhile, we see that {\MethodP} has the worst perplexity score.
However, it is important to note that even the ground truth tongue twisters score poorly here (around the same scores as {\MethodName} models). Tongue twisters do not usually use %are inherently less fluent in terms of utilizing
typical English tokens in standard sequences, thereby resulting in worse perplexity scores. 
Nonetheless, fluency/perplexity is still important as a sanity check for the basic qualities of the generated text (whether the generation is coherent, if it makes unnecessary grammatical errors, etc.). To this end, the fluency scores of {\MethodName} models are adequate.
% This shows that perplexity may not be the best measure of fluency for our task, and that fluency itself may not exactly correlate with the quality of a tongue twister (see \S\ref{sec:human_eval_results} for more).

Lastly, for keyword relevance, most scores are close to each other. The three {\MethodName} models have the three highest scores, indicating that {\MethodName} is able to generate difficult tongue twisters without compromising the task at hand.

\subsection{Human Evaluation Results}\label{sec:human_eval_results}
Table \ref{tab:results-human-full} shows the average results for the human evaluation. As with the automatic evaluations, we see an increase in phonetic difficulty when we introduce phonemes into the training process. Comparing g2g (GPT-2) and {\MethodJ} (GPT-2), we see an increase from 3.056 to 3.153 for {\TaskPrompt} and from 3.847 to 3.889 for {\TaskKeyword}. This trend also occurs for GPT-J models. Despite not being able to leverage existing GPT pretraining, {\MethodP} also works very well, outperforming all non-{\MethodName} models in all but one setting. These positive results indicate that incorporating phonetic information is indeed helpful.

In terms of phonetic difficulty, we observe that for {\TaskPrompt}, {\MethodP} works best for both GPT-2 and GPT-J, while for {\TaskKeyword}, {\MethodJ} works best for both GPT-2 and GPT-J. This may be because generating from keywords is generally more difficult than completing a prompt, so {\MethodJ} benefits from existing pretraining. Meanwhile, in terms of fluency, g2g methods work best for 2 settings, and {\MethodP} works best for 2 settings. Tongue twisters usually use words in creative and unnatural-sounding ways, and this may sometimes negatively affect fluency, so the "most fluent" sentence may not necessarily be the best tongue twister. Nevertheless, all the fluency metrics for {\MethodName} models are around 3.7 to 3.8, indicating good fluency. Overall, we conclude that {\MethodName} substantially improves phonetic difficulty while maintaining a competitive level of fluency.

\begin{table}[t]
    \scriptsize
    \centering
    \begin{tabular}{c|c|c}
        \textbf{Task} & \textbf{Pearson Correlation} & \textbf{Spearman Correlation} \\
        \hline
        {\TaskPrompt} & 0.117 ($p$=0.076) & 0.116 ($p$=0.079) \\
        {\TaskKeyword} & 0.102 ($p$=0.177) & 0.134 ($p$=0.075) \\
    \end{tabular}
    \caption{Corr. between human annotations and automatic metrics (BERT classifier) for phonetic difficulty.}
    \label{tab:correlations}
\end{table}

To verify the validity of using our BERT classifier to automatically measure phonetic difficulty, we compute correlations between the classifier's scores and human-annotated phonetic difficulty scores (Table \ref{tab:correlations}). We observe that the correlations are > 0.1 with relatively low p-values, which is an adequate score %reasonably good score
when comparing various automatic and human-annotated metrics \cite{gangal2021nareor}. While this can be further improved and is a limitation of our work, we also note that the automatic phonetic difficulty metric scores the gold outputs the highest (at 0.925) and the style transfer models lowest, which indicates that it is able to decently capture the notion of phonetic difficulty.

\begin{table}[t]
\centering
\small
\addtolength{\tabcolsep}{-4pt}
\resizebox{\columnwidth}{!}{
\begin{tabular}{|p{38pt}|p{253pt}|}
\hline 
\textbf{Method} & \textbf{Text} \\ \hline
\hline
Inputs & (ex.1) The pirates pirated ... \\ \hline
Gold out. & \textcolor{red}{The pirates pirated private property.} \\ \hline
g2g & \textcolor{brown}{The pirates pirated the pepper pot.} \\ \hline
Style T. & \textcolor{teal}{The pirates pirated more than \$50 million in online gambling in 2013.} \\ \hline 
PAN-P & \textcolor{violet}{The pirates pirated a little grape crepe. The pirate dread cratered a little.} \\ \hline
PAN-J & \textcolor{blue}{The pirates pirated the paired prince with a pair of paired pistols.
} \\ \hline
\hline
Inputs & (ex.2) If you stick ... \\ \hline
Gold out. & \textcolor{red}{If you stick a stock of liquor in your locker, it is slick to stick a lock upon your stock or some joker who is slicker is going to trick you of your liquor if you fail to lock your liquor with a lock?} \\ \hline
g2g & \textcolor{brown}{If you stick two sticks together, you get two sticks tied together.} \\ \hline
Style T. & \textcolor{teal}{If you stick with the same line, you have different things going for you.} \\ \hline 
PAN-P & \textcolor{violet}{If you stick a stock of land in a lamb and a lamb in a farm.} \\ \hline
PAN-J & \textcolor{blue}{If you stick a stock stick on the stock stick you stick, who's going to stick it on the stock stick you stick?} \\ \hline
\hline
\hline 
\textbf{Method} & \textbf{Text} \\ \hline \hline
Inputs & (ex.3) needles, noodles, kneadle, don, need \\ \hline
Gold out. & \textcolor{red}{I don't need your needles to kneadle noodles.} \\ \hline
g2g & \textcolor{brown}{One noodle needs a kneeler to lean on} \\ \hline
Style T. & \textcolor{teal}{Near the end of his reign, Leiblau had to send a delegate to Parliament, to support his friend King Gules of Northumbria.} \\ \hline 
PAN-P & \textcolor{violet}{If Donna Needles Need, who is Needles Nood, and the Need?} \\ \hline
PAN-J & \textcolor{blue}{Don’t kneadle a knuckle with a knuckle needle!} \\ \hline
\hline
Inputs & (ex.4) thieves, thankful, thanked, thirty, thursday\\ \hline
Gold out. & \textcolor{red}{The thirty-three thankful thieves thought that they thanked the other thirty-three thankful thieves throughout Thursday.} \\ \hline
g2g & \textcolor{brown}{The thieve thanked the thankful thief on Thursday.} \\ \hline
Style T. & \textcolor{teal}{Thanked by Thnx for the idea, I thought it was an idea that I wanted to do a spoof of Thankful Thankful and Thnx.} \\ \hline
PAN-P & \textcolor{violet}{Thankful thieves thought that they thought they thrilled the throne throughout Thursday.} \\ \hline
PAN-J & \textcolor{blue}{These sixty sheiks sent these thousand and sixty sheiks sixty sheiks thanking them for shouting these sixty sheiks sixty shouts on Thursday.} \\ \hline
\end{tabular}
}
\caption{\small Qualitative examples for both {\TaskPrompt} (first 2 examples) and {\TaskKeyword} (last 2 examples). We report only the best performing model based on phonetic difficulty from automatic evaluations for each type (in brackets): literal input, \textcolor{red}{gold output}, \textcolor{brown}{g2g (GPT-J)}, \textcolor{teal}{Style Transfer (BART)}, \textcolor{violet}{{\MethodP} (GPT-2+BART)}, and \textcolor{blue}{{\MethodJ} (GPT-J)}. Additional examples can be found in Appendix \ref{appendix:qual-examples}.
}
\label{tab:qualitative_examples}
\end{table}

\subsection{Qualitative Analysis}
\label{sec:qualitative-analysis}
Table \ref{tab:qualitative_examples} shows sample generations for both {\TaskPrompt} and {\TaskKeyword}. We observe that both PAN-P and PAN-J are able to use a wide variety of tongue twister techniques, such as rhyme (\textit{grape/crepe/crate-} in ex.1 PAN-P), 
alliteration (\textit{kneadle/knuckle} in ex.3 PAN-J), alternating final sounds (\textit{land/lamb} in ex.2 PAN-P), alternating initial sounds
(\textit{six-/sheik} in ex.4 PAN-J), and repetition. They are also able to generate proper nouns to suit the sentence, such as "\textit{Donna}" and "\textit{Needles Nood}" in ex.3 PAN-P. They can also combine multiple such techniques in a single tongue twister, such as \textit{stick/stock} and \textit{land/lamb} in ex.2 PAN-P.

Comparing this with the baseline methods (g2g and Style T.), we see that the generated outputs of the g2g baseline are decent and somewhat tongue twister-like but usually are very short and simple, often relying too much on alliterations. Meanwhile, the outputs for the style transfer methods are generally not tongue twisters. As discussed in \S\ref{subsec:auto-eval-results}, this is likely because it commonly fails at fully converting a regular sentence into a tongue twister.

For {\TaskPrompt}, we observe that even with a non-alliterative prompt such as "If you stick" (ex.2), the {\MethodName} models can still generate good tongue twisters, whereas the g2g method attempts to use repetition but the generated text is not that difficult to pronounce. Meanwhile, for the {\TaskKeyword} setting, PAN-J is able to incorporate the semantics of the words, rather than just copying the words themselves: in ex.4, PAN-J replaces "thirty/thieves" in the keywords with "sixty/sheiks". 
Lastly, comparing PAN-P and PAN-J, we see that PAN-J sentences generally sound smoother, while PAN-P sentences sometimes end rather abruptly ("\textit{in a farm.}" in ex.2; "\textit{and the Need?}" in ex.3). In addition, some of the outputs from PAN-P lack coherence (ex.3). This is likely because PAN-P uses a phoneme language model and hence is unable to leverage the large-scale pretraining from GPT models. On the other hand, this lack of large-scale pretraining can potentially free up PAN-P to use more diverse tongue twister techniques, such as rhymes (ex.1) and proper nouns (ex.3) which are less common in PAN-J.

\section{Related Work}
\label{sec:related_work}
Automatic tongue twister generation is a largely unexplored task. Existing systems mostly use synonym replacements \cite{zeng_2019} on existing tongue twisters, which requires a large list of tongue twisters to begin with and cannot generate novel ones from scratch. \citet{carey_tt} generates tongue twisters using sound vectors, and \citet{kaggle_tt} trains an LSTM on a small tongue twister dataset, but neither are able to produce novel and semantically coherent examples. Furthermore, no methods currently exist for the {\TaskKeyword} task. 

There have been multiple studies on creative generation of various figures of speech such as similes \cite{chakrabarty-etal-2020-generating}, metaphors \cite{chakrabarty-etal-2021-mermaid}, and personifications \cite{keh-etal-2022-pineapple}. However, these other creative linguistic constructs don't require working with another modality in the same way that tongue twister generation relies on phonemes. Among these creative linguistic constructs, the closest ones to tongue twisters would likely be alliterations \cite{hopkins-kiela-2017-automatically}, rhymes \cite{xue-etal-2021-deeprapper}, and poetry \cite{hafez, Cruys2020AutomaticPG}. Notably, DeepHaiku \cite{haiku} also explores using phonemes and a multitask objective in order to generate haikus. However, tongue twister generation goes beyond alliterations and rhymes; rather it is a mix of all these various techniques. In addition, it differs from poetry generation because poetry generation focuses on generating rhythmic verses and syllables, whereas the main focus of tongue twister generation is on phonetic difficulty.

Using phonemes in language modeling has been previously explored in the speech domain for automatic speech recognition \cite{sundararaman2021phoneme, Belinkov2019AnalyzingPA, Xu2021SimpleAE}. In this paper, we trained a BART model to do p2g translation. Other existing methods include expectation maximization \cite{knight-etal-2006-unsupervised}, 
A* search \cite{corlett-penn-2010-exact}, and Hidden Markov Models \cite{hopkins-kiela-2017-automatically}.

There is also work on more general constrained text generation tasks. An example is \citet{feng-etal-2019-keep}, who propose Semantic Text Exchange to adjust a text's topic-level semantics. \citet{lin-etal-2020-commongen} introduce a generative commonsense reasoning task using keyword-to-text generation called CommonGen. %Works investigating this task include EKI-BART \cite{fan2020enhanced} and KG-BART \cite{liu2020kg}, which use external knowledge to enhance performance on CommonGen.
SAPPHIRE \cite{feng-etal-2021-sapphire} and VisCTG \cite{feng_caption} investigate approaches to improve performance on CommonGen, the latter using per-example visual grounding.

\section{Conclusion and Future Work}
\label{sec:conclusion_future_work}
In this paper, we proposed the task of automatic tongue twister generation, and explored it under two settings: {\TaskPrompt} and {\TaskKeyword}. We curated a dataset called {\DatasetName} of 600+ English tongue twisters from various sources and proposed {\MethodName}, a training methodology which incorporates phoneme representations. We implemented two variants: {\MethodP} (Phoneme), which trains a phoneme-based language model, and {\MethodJ} (Joint), which jointly incorporates both phoneme-level information and grapheme-level information during training time. Through empirical results and qualitative evaluations, we showed that incorporating phonemes is indeed helpful in producing effective tongue twisters which are harder to pronounce while staying fluent. 

While {\MethodName} works well at generation, the generation process lacks interpretability. This is most notable when looking at the phonetic difficulty classifier. 
Currently, the classifier does not identify and separately score elements of phonetic difficulty or come up with an explicit decomposition. We believe that such explicit decomposition can be very useful in the future for understanding more about tongue twisters and the "mouth model" discussed in \S\ref{sec:intro}. In addition, the procedure of fine-tuning a pretrained model in 4 different modes involving both graphemes and phonemes, devised here to fine-tune {\MethodJ}, can also be adopted more generally as a data augmentation \cite{feng-etal-2021-survey,feng-etal-2020-genaug} method for LM fine-tuning, creating a (pseudo) count of 4N training examples given N initial ones. Lastly, tongue twisters can potentially be incorporated in dialogue agents, adding creativity and personality \cite{Li_Jiang_Feng_Sprague_Zhou_Hoey_2020}.

% \clearpage
\section*{Limitations}
As mentioned in \S \ref{sec:conclusion_future_work}, our current model and classifer are deficient in terms of their interpretability on certain aspects, and would greatly benefit from addition to their interpretability on these fronts.

Our models and datasets are limited to English tongue twisters \cite{bender2018data}. In addition, when we convert to the phoneme space, we only use IPA phonemes. We selected IPA because it is the most common and most readily available phonemization method. However, there also exist many other phonemization methods such as the ARPAbet / CMU Pronouncing Dictionary \cite{cmudict}, the SAMPA \cite{sampa}, or the Festival phonemization scheme \cite{festival}. We also choose to use deep-phonemizer for g2p transliteration. There are many other available phonemization tools such as Epitran \cite{Mortensen-et-al:2018} and the \texttt{phonemizer} Python package \cite{Bernard2021}. It would be interesting to explore how the performance will change if we try other phoneme alphabets or phonemization methods.

In addition, we see in \S \ref{sec:results_and_analysis} that the style transfer models do not really work that well. In this paper, we only tried simple BART and T5 seq2seq models. One possible way to expand on this would be to try other more sophisticated style transfer methods.

Due to computational resources, we were not able to explore larger models and had to use a compressed version of GPT-J, which may have slightly affected the performance.

\section*{Ethics}

The {\DatasetName} dataset we propose and release herein has undergone a per-example, manual, vetting process during its curation and pre-processing stage, as described further in Appendix \ref{appendix:dataset-vetting}, which removes examples which may exhibit offensive words, profanities, racism, gender bias, and other malicious language.

We do collect human evaluation ratings using crowd-sourcing, specifically through AMT. However, we neither solicit, record, nor request any kind of personal or identity information from the annotators. Our AMT annotation was conducted in a manner consistent with terms of use of any sources and intellectual property and privacy rights of AMT crowd workers. Crowdworkers were fairly compensated: \$0.56 per fluency and phonetic difficulty HIT, for roughly 2 min tasks. This is at least 2 times the minimum U.S.A. wage of \$7.25 per hour (\$0.56 per 2 mins is around \$16.8 per hour).

NLG models are known to suffer from biases learnable from training or finetuning on data, such as gender bias \cite{dinan2020queens}. However, our work and contribution does not present or release any completely new model architectures, and is primarily concerned with more careful adaptation and finetuning of existing pretrained models for a particular class of creative linguistic constructs (i.e. tongue twisters). The frailties, vulnerabilities, and potential dangers of these models have been well researched and documented, and a specific re-investigation would be repetitive and beyond the scope and space constraints of this paper.

We do not foresee any explicit way that malicious actors could specifically misuse finetuned models that could be trained on our data, beyond the well-researched, aforementioned misuse that is possible in general with their instantiation for any transduction task or dataset (e.g. summarization).
%\section*{Acknowledgements}

\bibliography{anthology,custom}
\bibliographystyle{acl_natbib}

%\newpage
\bigskip
\appendix
\section{Additional Details --- Dataset Collection}\label{appendix:dataset-details}
\subsection{Sources}
\label{appendix:dataset-sources}
We curate our tongue twisters from a heterogeneous mix of online sources, including but not limited to the ones listed below:
\begin{enumerate}
    \item \href{https://service.uark.edu/mentoring_programs/vacliteracy/pdf/english_tongue_twisters.pdf}{University of Arkansas}
    \item \href{https://www.reddit.com/r/tonguetwisters/}{The r/tonguetwister Subreddit}
       \item \href{https://www.reddit.com/r/AskReddit/comments/20qxz5/what_is_your_favorite_tongue_twister/}{Various AskReddit threads}
    \item \href{https://www.mondly.com/blog/2019/08/23/71-best-tongue-twisters-to-perfect-your-english-pronunciation/}{Mondly.com}
    \item \href{http://www.uebersetzung.at/twister/en.htm}{Uebersetzung}
    \item \href{https://books.google.com/books?id=30x4CwAAQBAJ&pg=PP1&lpg=PP1&dq=tongue+twisters+collection&source=bl&ots=lMvYXs98wn&sig=lq_bwqTOMsX8CgTZG8PqREbYSZg&hl=en&sa=X&ved=0ahUKEwjn-r_m06zVAhWEFz4KHVdfDTQ4ChDoAQgnMAA#v=onepage&q=tongue\%20twisters}{Marcus Stuart's LOL Tongue Twisters book}
    \item \href{https://languageavenue.com/teachers/teaching-ideas/english-tongue-twisters}{Language Avenue}
    \item \href{http://bilingualmonkeys.com/22-funny-tongue-twisters-for-kids/}{Bilingual Monkeys}
    \item \href{http://pun.me/pages/tongue-twisters.php}{Pun.me}
    \item \href{http://www.esl4kids.net/tongue.html}{ESL}
    \item \href{http://www.sweetrhymes.com/tongue-twisters/alphabetical-collection-of-tongue-twisters/}{Sweetrhymes}
    \item \href{https://www.engvid.com/english-resource/50-tongue-twisters-improve-pronunciation/}{EngVid}
    \item \href{https://www.engvid.com/english-resource/50-tongue-twisters-improve-pronunciation/}{IvyPanda}
\end{enumerate}

\subsection{Vetting}
\label{appendix:dataset-vetting}
We then perform the following filtering steps to retain only a collection of high-quality dataset examples:
\begin{itemize}
    \item Remove near-repetitive examples to ensure each example is unique
    \item Remove excessively short or meaningless examples lacking sentence structure, e.g. \textit{blue blood, bad blood}
    \item Remove poems or rhymes
    \item Remove examples containing offensive words, racism, gender bias or other harmful and malicious language of any nature, to prevent models learnt from this data from further ingraining or amplifying such phenomena.
\end{itemize}

\section{Appendix B: Evaluation Details}\label{appendix:human-eval}

% \subsection{Human Evaluation Setup}
% \label{appendix:human-eval-setup}
To prevent annotator judgements for one attribute from inadvertently influencing the other, we conduct the studies for soliciting Fluency and Phonetic Difficulty scores separately.

Averaging over the 4 settings described in \S\ref{subsec:human-eval} ( TT-Keyword/TT-Prompt $\times$ GPT-2/GPT-J), a total of 20 unique AMT annotators participated in the study for Fluency, each performing 3.6 HITs on average. Annotators were compensated \$0.56 per HIT, each of which was designed to take < 2 mins on average.

Averaging over the 4 settings, 16.51 unique AMT annotators participated in the second, separate study for Phonetic Difficulty, each performing 4.36 HITs on average. Annotators were compensated \$0.56 per HIT, each of which was designed to take < 2 mins on average.

%The html template including instructions, questions, and other details can be found in 
%a file named \texttt{template.html} in our code submission zip. 

% \subsection{Inter-Annotator Agreement (IAA) Scores}
% \label{appendix:inter-annotator-agreement-scores}

% \begin{table}[t]
%     \small
%     \centering
%     \begin{tabular}{c|c|c}
%          \textbf{Metric} &  \begin{tabular}[t]{@{}c@{}} \textbf{Spearman} \\ \textbf{Correlation} \end{tabular} & \textbf{Krippendorff $\alpha$} \\
%          \hline
%          Phonetic Difficulty & 0.071 & 0.092 \\
%          Fluency & 0.006 & 0.043\\
%     \end{tabular}
%     \caption{Inter-annotator agreement scores.}
%     \label{tab:inter-annotator-scores}
% \end{table}

% See Table \ref{tab:inter-annotator-scores} for IAA scores. To get the Spearman correlation point value for a given aspect and test instance, we compute mean pairwise Spearman correlation between the aspect values assigned to the corresponding model outputs by every pair of annotators. Specifically, we use the \textit{scipy.stats} implementation to compute this.\footnote{\url{https://docs.scipy.org/doc/scipy/reference/generated/scipy.stats.spearmanr.html}}

% For Krippendorff $\alpha$, we treat each human evaluation aspect as an ordinal quantity. Specifically, we use the implementation provided by the python library \textit{krippendorff 0.5.1}.\footnote{\url{https://pypi.org/project/krippendorff/}}

\section{Further Implementation Details}
\label{appendix:hyperparameter-settings}
In \S \ref{sec:implementation-settings}, we detailed the hyperparameters used for pre-training BART, as well as for fine-tuning GPT-2, GPT-J, BART, and T5. 
We conduct a hyperparameter search to check which values led to the best performance. For learning rate, we tried \{1e-6, 5e-6, 1e-5, 2e-5, 2e-5, 1e-4\}; for batch size, we tried \{2,4,8,16\}; and for number of epochs, we tried \{2, 5, 10, 20\}. These search bounds were selected based on known commonly-used values for these models. We start with a baseline model of lr=2e-5, bsz=8 and 10 epochs, and individually change each setting to investigate its effect on performance. One trial was conducted per hyperparameter setting. We use a maximum sequence length of 256. In terms of other hyperparameters, we mostly used default values which are known to work for these models. This includes the warmup steps and learning rate decays, which we detail in \S \ref{sec:implementation-settings}. (Note: the above hyperparameter search settings are for fine-tuning. We could not do an extensive hyperparameter search for pre-trainig due to time constraints. We ran pre-training twice to test the effect of learning rate 1e-4 vs. 5e-4, and ultimately selected 5e-4.)

Model selection was done based on the epoch with the best validation loss. We report the best validation losses for each training process below: GPT-2 pre-training=0.77, GPT-2 fine-tuning ({\TaskPrompt})=1.08, GPT-2 fine-tuning ({\TaskKeyword}) =0.95, GPT-J fine-tuning ({\TaskPrompt})=1.23, GPT-J fine-tuning ({\TaskKeyword})=1.19, BART-fine-tuning ({\TaskKeyword}) = 1.31, BART-fine-tuning ({\TaskPrompt}) = 1.30, T5-fine-tuning ({\TaskKeyword}) = 1.45, T5-fine-tuning ({\TaskPrompt}) = 1.52.

All training was done on Google Colaboratory environments using V100 GPUs with 16GB memory. Pre-training GPT-2 for the p2p model took 90 minutes per epoch, and pre-training BART for the p2g model took 70 minutes per epoch. In terms of fine-tuning on {\DatasetName}, GPT-2 takes around 1 minute per epoch, GPT-J around 2 minutes per epoch, BART around 1 minute per epoch, and T5 around 1 minute per epoch.

\section{Additional Qualitative Examples}
\label{appendix:qual-examples}

In \S \ref{sec:qualitative-analysis}, we only report the best performing models for each of the methods. In Table \ref{tab:extra_qualitative_examples}, we extend this to all models for all the methods. We also provide a few more examples for both {\TaskPrompt} and {\TaskKeyword}.

\begin{table}[t]
\centering
\small
\addtolength{\tabcolsep}{-4pt}
\resizebox{\columnwidth}{!}{
\begin{tabular}{|p{62pt}|p{252pt}|}
\hline 
\textbf{Method} & \textbf{Text} \\ \hline
\hline
Inputs & (ex.1) The pirates pirated ... \\ \hline
Gold out. & \textcolor{red}{The pirates pirated private property.} \\ \hline
g2g (GPT2) & \textcolor{purple}{The pirates pirated the property.} \\ \hline
g2g (GPTJ) & \textcolor{brown}{The pirates pirated the pepper pot.} \\ \hline
Style T. (BART) & \textcolor{teal}{The pirates pirated more than \$50 million in online gambling in 2013.} \\ \hline 
Style T. (T5) & \textcolor{olive}{The pirates pirated more than \$50 million in online gambling in 2013, according to a government-backed study.} \\ \hline 
PAN-P & \textcolor{violet}{The pirates pirated a little grape crepe. The pirate dread cratered a little.} \\ \hline
PAN-J (GPT2) & \textcolor{darkgray}{The pirates pirated the music and the Roman river carried the criminals off.} \\ \hline
PAN-J (GPTJ) & \textcolor{blue}{The pirates pirated the paired prince with a pair of paired pistols.} \\ \hline
\hline
Inputs & (ex.2) If you stick ... \\ \hline
Gold out. & \textcolor{red}{If you stick a stock of liquor in your locker, it is slick to stick a lock upon your stock or some joker who is slicker is going to trick you of your liquor if you fail to lock your liquor with a lock?} \\ \hline
g2g (GPT2) & \textcolor{purple}{If you stick a weight silky liquor on your wrist, you should stick a weight silky liquor on your inner wrist.} \\ \hline
g2g (GPTJ) & \textcolor{brown}{If you stick two sticks together, you get two sticks tied together.} \\ \hline
Style T. (BART) & \textcolor{teal}{If you stick with the same line, you have different things going for you.} \\ \hline 
Style T. (T5) & \textcolor{olive}{If you stick with the same line, you have different things going for yout.} \\ \hline 
PAN-P & \textcolor{violet}{If you stick a stock of land in a lamb and a lamb in a farm.} \\ \hline
PAN-J (GPT2) & \textcolor{darkgray}{If you stick a stock of liquor in your locker it is slick to put a lock on your stock.} \\ \hline
PAN-J (GPTJ) & \textcolor{blue}{If you stick a stock stick on the stock stick you stick, who's going to stick it on the stock stick you stick?} \\ \hline
\hline
Inputs & (ex.3) I saw Susie ... \\ \hline
Gold out. & \textcolor{red}{I saw Susie sitting in a shoe shine shop.} \\ \hline
g2g (GPT2) & \textcolor{purple}{I saw Susie sitting on the sofa with a big black basket of biscuits.} \\ \hline
g2g (GPTJ) & \textcolor{brown}{I saw Susie sitting on a buttercup.} \\ \hline
Style T. (BART) & \textcolor{teal}{I saw Susie and then she came and took me to the kitchen where we cleaned.} \\ \hline 
Style T. (T5) & \textcolor{olive}{I saw Susie and then she came and took me to the kitchen where we cleaned up and we started talking about the new dog," she recalled.} \\ \hline 
PAN-P & \textcolor{violet}{I saw Susie sitting on a seasawe horse. and I saw haunted houses.} \\ \hline
PAN-J (GPT2) & \textcolor{darkgray}{I saw Susie sitting in a speck of sheared sheep's sheep's sheep's sheep skin.} \\ \hline
PAN-J (GPTJ) & \textcolor{blue}{I saw Susie sitting on a butter bucket with a black butter bucket sat on her butter bucket.} \\ \hline

\hline 
\textbf{Method} & \textbf{Text} \\ \hline \hline
Inputs & (ex.4) needles, noodles, kneadle, don, need \\ \hline
Gold out. & \textcolor{red}{I don't need your needles to kneadle noodles.} \\ \hline
g2g (GPT2) & \textcolor{purple}{Don's dirt feet meet your earrings at the bottom of the dirt cany bore trail.} \\ \hline
g2g (GPTJ) & \textcolor{brown}{One noodle needs a kneeler to lean on} \\ \hline
Style T. (BART) & \textcolor{teal}{Near the end of his reign, Leiblau had to send a delegate to Parliament, to support his friend King Gules of Northumbria.} \\ \hline 
Style T. (T5) & \textcolor{olive}{Near the end of his reign, Leiblau had to send a delegate to Parliament, to support his friend King Gules of Northumbria.} \\ \hline 
PAN-P & \textcolor{violet}{If Donna Needles Need, who is Needles Nood, and the Need?} \\ \hline
PAN-J (GPT2) & \textcolor{darkgray}{If you crave a soft chew toy, buy a chew toy that needs to chew.} \\ \hline
PAN-J (GPTJ) & \textcolor{blue}{Don’t kneadle a knuckle with a knuckle needle!} \\ \hline
\hline

Inputs & (ex.5) thieves, thankful, thanked, thirty, thursday\\ \hline
Gold out. & \textcolor{red}{The thirty-three thankful thieves thought that they thanked the other thirty-three thankful thieves throughout Thursday.} \\ \hline
g2g (GPT2) & \textcolor{purple}{Twelve thieves today took sixty dollars from Thies' birthday party.} \\ \hline
g2g (GPTJ) & \textcolor{brown}{The thieve thanked the thankful thief on Thursday.} \\ \hline
Style T. (BART) & \textcolor{teal}{Thanked by Thnx for the idea, I thought it was an idea that I wanted to do a spoof of Thankful Thankful and Thnx.} \\ \hline 
Style T. (T5) & \textcolor{olive}{Thanked by Thnx for the idea, I thought it was an idea that I wanted to do a spoof of Thankful Thankful and Thnx, so it went to the video website, Twitter, and did some research on a T shirt.} \\ \hline 
PAN-P & \textcolor{violet}{Thankful thieves thought that they thought they thrilled the throne throughout Thursday.} \\ \hline
PAN-J (GPT2) & \textcolor{darkgray}{I'm grateful tonight for thanking the valiant brave thieves.} \\ \hline
PAN-J (GPTJ) & \textcolor{blue}{These sixty sheiks sent these thousand and sixty sheiks sixty sheiks thanking them for shouting these sixty sheiks sixty shouts on Thursday.} \\ \hline

Inputs & (ex.6) snails, silently, sailed, slimy \\ \hline
Gold out. & \textcolor{red}{Six slimy snails sailed silently.} \\ \hline
g2g (GPT2) & \textcolor{purple}{Six silently sullen sullen snails slid silently across the sea.} \\ \hline
g2g (GPTJ) & \textcolor{brown}{Slimy snails sailed silently.} \\ \hline
Style T. (BART) & \textcolor{teal}{The ship was spotted by U.S. Coast Guard helicopter from a distance of four thousand, seven hundred nautical miles.} \\ \hline 
Style T. (T5) & \textcolor{olive}{The ship was spotted by Uas was was was was by by by by by by by.} \\ \hline 
PAN-P & \textcolor{violet}{Sailed from static line seven and all silently from Trondheim at seven on Sunday night.} \\ \hline
PAN-J (GPT2) & \textcolor{darkgray}{Sailing silently on the sleigh.} \\ \hline
PAN-J (GPTJ) & \textcolor{blue}{Six squandered snails silently sailed in a slivy ship.} \\ \hline
\hline
\end{tabular}
}
\caption{\small Additional qualitative examples for both {\TaskPrompt} (first 3) and {\TaskKeyword} (last 3): literal input, \textcolor{red}{gold output}, \textcolor{purple}{g2g (GPT-2)}, \textcolor{brown}{g2g (GPT-J)}, \textcolor{teal}{Style Transfer (BART)}, \textcolor{olive}{Style Transfer (T5)}, \textcolor{violet}{{\MethodP} (GPT-2+BART)}, \textcolor{darkgray}{{\MethodJ} (GPT-2)}, and \textcolor{blue}{{\MethodJ} (GPT-J)}.}
\label{tab:extra_qualitative_examples}
\end{table}

\end{document}